\documentclass[letterpaper, 10 pt, conference]{ieeeconf}  
\IEEEoverridecommandlockouts
\usepackage{graphics} % for pdf, bitmapped graphics files
\usepackage{epsfig} % for postscript graphics files
\usepackage{mathptmx} % assumes new font selection scheme installed
\usepackage{times} % assumes new font selection scheme installed
\usepackage{amsmath} % assumes amsmath package installed
\usepackage{amssymb}  % assumes amsmath package installed
\usepackage{graphicx}
\usepackage{amsmath,amssymb} % define this before the line numbering.
\usepackage{color}
\usepackage[ruled,vlined]{algorithm2e}
\usepackage{makecell}
\usepackage[utf8]{inputenc} % allow utf-8 input
\usepackage[T1]{fontenc}    % use 8-bit T1 fonts
\usepackage{hyperref}       % hyperlinks
\usepackage{url}            % simple URL typesetting
\usepackage{booktabs}       % professional-quality tables
\usepackage{amsfonts}       % blackboard math symbols
\usepackage{nicefrac}       % compact symbols for 1/2, etc.
\usepackage{microtype}      % microtypography
\usepackage{subcaption}

\title{\LARGE \bf
Risk Averse Robust Adversarial Reinforcement Learning
}

\author{% <-this % stops a space
Xinlei Pan$^1$, Daniel Seita$^1$, Yang Gao$^1$, John Canny$^1$
\thanks{$^1$ University of California, Berkeley.
{\scriptsize\{xinleipan,seita,yg,canny\}@berkeley.edu}
}% <-this % stops a space
\thanks{
        {}}%
\thanks{}%
}

\begin{document}
\maketitle

\begin{abstract}
Deep reinforcement learning has recently made significant progress in solving computer games and robotic control tasks. A known problem, though, is that policies overfit to the training environment and may not avoid rare, catastrophic events such as automotive accidents. A classical technique for improving the robustness of reinforcement learning algorithms is to train on a set of randomized environments, but this approach only guards against common situations. Recently, robust adversarial reinforcement learning (RARL) was developed, which allows efficient applications of random and systematic perturbations by a trained adversary. A limitation of RARL is that only the expected control objective is optimized; there is no explicit modeling or optimization of risk. Thus the agents do not consider the probability of catastrophic events (i.e., those inducing abnormally large negative reward), except through their effect on the expected objective. In this paper we introduce risk-averse robust adversarial reinforcement learning (RARARL), using a risk-averse protagonist and a risk-seeking adversary. We test our approach on a self-driving vehicle controller. We use an ensemble of policy networks to model risk as the variance of value functions. We show through experiments that a risk-averse agent is better equipped to handle a risk-seeking adversary, and experiences substantially fewer crashes compared to agents trained without an adversary.  
Supplementary materials are available at \url{https://sites.google.com/view/rararl}.
\end{abstract}


\section{Introduction}
Reinforcement learning has demonstrated remarkable performance on a variety of sequential decision making tasks such as Go \cite{silver2016mastering}, Atari games \cite{mnih2015human}, autonomous driving \cite{shalev2016safe,you2017virtual}, and continuous
robotic control \cite{lillicrap2015continuous,benchmarking_2016}. Reinforcement learning (RL) methods fall under two broad categories: model-free and model-based. In model-free RL, the environment's physics are not modeled, and such methods require substantial environment interaction and can have prohibitive sample complexity~\cite{trpo}. In contrast, model-based methods allow for systematic analysis of environment physics, and in principle should lead to better sample complexity and more robust policies.  These methods, however, have to date been challenging to integrate with deep neural networks and to generalize across multiple environment dimensions~\cite{zhu_2018,model_free_based_2018}, or in truly novel scenarios, which are expected in unrestricted real-world applications such as driving.

\begin{figure}[t!]
\centering
\includegraphics[width=0.8\linewidth]{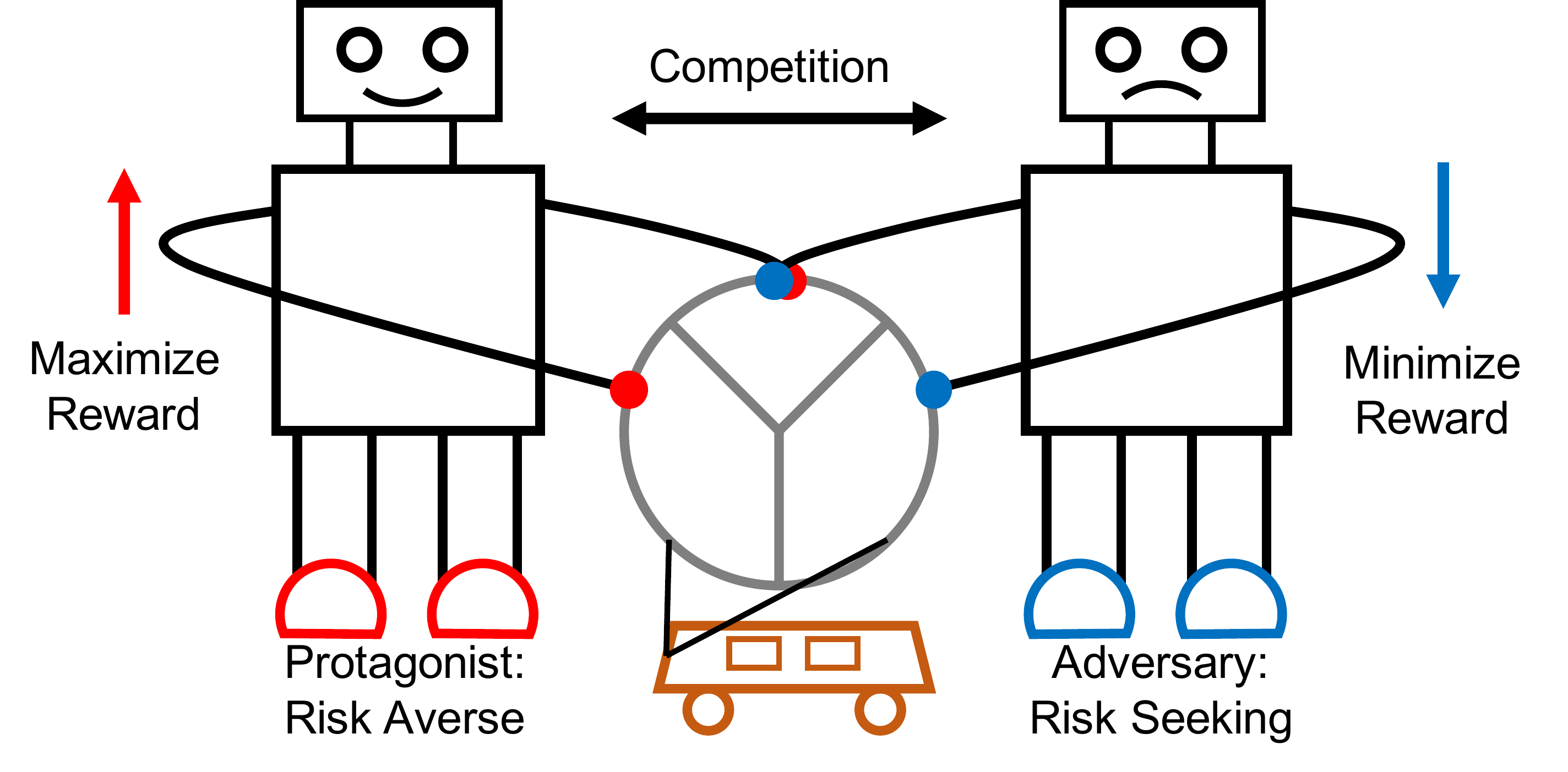}
\caption{
\footnotesize
Risk averse robust adversarial reinforcement learning diagram: an autonomous
driving example. Our framework includes 
two competing agents acting against each other, 
trying to drive a car (protagonist), or trying to slow or crash the car (adversary). We include a notion of risk modeling in policy learning. The risk-averse protagonist and risk-seeking adversarial agents learn policies
to maximize or minimize reward, respectively. The use of the adversary helps the protagonist to effectively explore risky states.
}
\label{fig1}
\vspace*{-10pt}
\end{figure}

In this work, we focus on model-free methods, but include \emph{explicit modeling of risk}. We additionally focus on a framework that includes an \emph{adversary} in addition to the main (i.e., protagonist) agent. By modeling risk, we can train stronger adversaries and through competition, more robust policies for the protagonist (see Figure~\ref{fig1} for an overview). We envision this as enabling training of more robust agents in simulation and then using sim-to-real techniques~\cite{sadeghi2016cad2rl} to generalize to real world applications, such as house-hold robots or autonomous driving, with high reliability and safety requirements.
A recent algorithm combining robustness in reinforcement learning and the adversarial framework is robust adversarial reinforcement learning (RARL)~\cite{pinto2017robust}, which trained a robust protagonist agent by having an adversary providing random and systematic attacks on input states and dynamics. The adversary is itself trained using reinforcement learning, and tries to minimize the long term expected reward while the protagonist tries to \emph{maximize} it. As the adversary gets stronger, the protagonist experiences harder challenges. 

RARL, along with similar methods~\cite{mandlekaradversarially}, is able to achieve some robustness, but the level of variation seen during training may not be diverse enough to resemble the variety encountered in the real-world. Specifically, the adversary does not actively seek catastrophic outcomes as does the agent constructed in this paper. Without such experiences, the protagonist agent will not learn to guard against them.
 Consider autonomous driving: a car controlled by the protagonist may suddenly be hit by another car. We call this and other similar events \emph{catastrophic} since they present extremely negative rewards to the protagonist, and should not occur under a reasonable policy. Such catastrophic events are highly unlikely to be encountered if
 an adversary only randomly perturbs the environment parameters or dynamics, or if the adversary only tries to minimize total reward.

In this paper, we propose risk averse robust adversarial reinforcement learning (RARARL) for training risk averse policies that are simultaneously robust to dynamics changes. Inspired by~\cite{tamar2016learning}, we model risk as the variance of value functions. To emphasize that the protagonist be averse to catastrophes, we design an asymmetric reward function (see Section~\ref{ssec:sim_env}): successful behavior receives a small positive reward, whereas catastrophes receive a  very negative reward.

A robust policy should not only maximize long term expected reward, but should also select actions with low variance of that expected reward. Maximizing the expectation of the value function only maximizes the point estimate of that function without giving a guarantee on the variance. While~\cite{tamar2016learning} proposed a method to estimate that variance, it assumes that the number of states is limited, while we don't assume limited number of states and that assumption makes it impractical to apply it to real world settings where the number of possible states could be infinitely large. Here, we use an ensemble of Q-value networks to estimate variance. A similar technique was proposed in Bootstrapped DQNs~\cite{osband2016deep} to assist exploration, though in our case, the primary purpose of the ensemble is to estimate variance.

We consider a two-agent reinforcement learning scenario (formalized in Section~\ref{sec:formalism}). Unlike in~\cite{pinto2017robust}, where the agents performed actions simultaneously, here they take turns executing actions, so that one agent may take multiple steps to bring the environment in a more challenging state for the other. We seek to enable the adversarial agent to actively explore the parameter variation space, so that the perturbations are generated more efficiently. We use a discrete control task, autonomous driving with the TORCS~\cite{wymann2000torcs} simulator, to demonstrate the benefits of RARARL.
\section{Related Work}

\textbf{Reinforcement Learning with Adversaries}. A recent technique in reinforcement learning involves introducing adversaries and other agents that can adjust the environment difficulty for a main agent. This has been used for robust grasping~\cite{supervision_competition_2017}, simulated fighting~\cite{multi_agent_2018}, and RARL~\cite{pinto2017robust}, the most relevant prior work to ours. RARL trains an adversary to appropriately perturb the environment for a main agent. The perturbations, though, were limited to a few parameters such as mass or friction, and the trained protagonist may be vulnerable to other variations.

The works of~\cite{mandlekaradversarially} and~\cite{pattanaik2017robust} proposed to add 
noise to state observations to provide adversarial perturbations, with the noise generated using fast gradient sign method~\cite{goodfellow2014explaining}. However, they did not consider training an adversary
or training risk averse policies.
The work of~\cite{paul2018alternating} proposed to introduce Bayesian optimization to actively select environment variables that may induce catastrophes, so that  models trained can be robust to these environment dynamics.  However, they did not systematically explore dynamics variations and therefore the model may be vulnerable to changing dynamics even if it is robust to a handful of rare events. 

\textbf{Robustness and Safety in RL}. More generally, robustness and safety have long been explored in reinforcement learning~\cite{risk_sensitive_1998,risk_aversion_2017,percentile_risk_2018}. 
Chow et al.~\cite{percentile_risk_2018} proposed to model risk via constraint
or chance constraint on the conditional value at risk (CVaR). This paper provided strong
convergence guarantees but made strong assumptions: value and constrained value functions are assumed to
be known exactly and to be differentiable and smooth. Risk is estimated by simply sampling trajectories which may never encounter adverse outcomes, whereas with sparse risks (as is the case here) adversarial sampling provides more accurate estimates of the probability of a catastrophe.

A popular ingredient is to enforce \emph{constraints} on an agent during exploration~\cite{safe_exploration_2012} and policy updates~\cite{cpo_2017,prob_safe_transfer_2017}. Alternative techniques include random noise injection during various stages of training~\cite{parameter_noise_2018,noisy_nets_2018}, 
injecting noise to the transition dynamics during
training~\cite{rajeswaran2016epopt}, learning when to reset~\cite{learning_to_reset_2018} and even physically crashing as needed~\cite{crashing2017}. However, Rajeswaran et al.~\cite{rajeswaran2016epopt}
requires training on a target domain and experienced
performance degradation when the target domain has a different model parameter distribution from the source. 
% Previous work~\cite{rajeswaran2016epopt} shows that by sampling from a source distribution of limited model parameters, their model achieves superior performance when its parameter varies. However, their results indicate performance degradation when the target domain has a different model parameter distribution from the source. 
% Their method also only samples a limited number of parameters in a hand-engineered manner.
We also 
note that in control theory,~\cite{aswani2013provably, aswani2012extensions} have provided theoretical analysis for robust control, though their
focus lies in model based RL instead of model free RL.
These prior techniques are orthogonal to our contribution, which relies on \emph{model ensembles} to estimate variance. 

\textbf{Uncertainty-Driven Exploration}. Prior work on exploration includes~\cite{pathak2017curiosity}, which measures novelty of states using state prediction error, and~\cite{bellemare2016unifying}, which uses pseudo counts to explore novel states. In our work, we seek to measure the risk of a state by the variance of value functions. The adversarial agent explores states with high variance so that it can create appropriate challenges for the protagonist.

\textbf{Simulation to Real Transfer}. Running reinforcement learning on physical hardware can be dangerous due to exploration and slow due to high sample complexity. One approach to deploying RL-trained agents safely in the real world is to experience enough environment variation during training in simulation so that the real-world 
environment looks just like another variation. These simulation-to-real techniques have grown popular, including domain randomization~\cite{sadeghi2016cad2rl, tobin2017domain} and dynamics randomization~\cite{peng2017sim}. However, their focus is on transferring policies to the real world rather than training
robust and risk averse policies.
% \textbf{Reinforcement Learning for Autonomous Driving}. 
% Reinforcement learning is a promising direction for training autonomous
% vehicles \cite{you2017virtual,gao2018reinforcement,li2017infogail}. 
% Vehicles trained using reinforcement learning should be robust and 
% risk averse before they can be deployed for daily use. To our 
% knowledge, here we provide the first study of training risk 
% averse and robust policies for simulated autonomous driving.
 
\section{Risk Averse Robust Adversarial RL}\label{sec:formalism}
In this section, we formalize our risk averse robust adversarial reinforcement learning (RARARL) framework.

\subsection{Two Player Reinforcement Learning}
We consider the environment as a Markov Decision Process (MDP) $\mathcal{M} = \{\mathcal{S},\mathcal{A},\mathcal{R},\mathcal{P},\gamma\}$,
where $\mathcal{S}$ defines the state space, $\mathcal{A}$ defines the action
space, $\mathcal{R}(s,a)$ is the reward function, $\mathcal{P}(s'|s,a)$ 
is the state transition model, and $\gamma$ is the reward discount rate.
There are two agents: the {\em protagonist}
$P$ and the {\em adversary} $A$. 

\textbf{\textit{Definition.}} Protagonist Agent. A protagonist $P$ learns a policy $\pi_P$ to maximize discounted expected
reward $\mathbb{E}_{\pi_P}[\sum{\gamma^t}r_t]$. The protagonist should
be risk averse, so we define the value of action $a$ at state $s$ to be
\begin{equation}
\hat{Q}_P(s,a) = Q_P(s,a) - \lambda_P Var_{k}[Q^k_P(s,a)],
\label{eq:computeq1}
\end{equation}
where $\hat{Q}_P(s,a)$ is the modified Q function, $Q_P(s,a)$ is the 
original Q function, and $Var_{k}[Q^k_P(s,a)]$ is the variance of the Q function
across $k$ different models, and $\lambda_P$ is a constant; 
The term $-\lambda_P Var_{k}[Q^k_P(s,a)]$ is called
the \emph{risk-averse} term thereafter, and encourages the protagonist to seek lower variance actions. The reward for $P$ is the environment reward $r_t$ at time $t$. 

\textbf{\textit{Definition.}} Adversarial Agent. An adversary $A$ learns a policy $\pi_A$ to \emph{minimize} long term expected
reward, or to maximize the negative discounted reward 
$\mathbb{E}_{\pi_A}[\sum-{\gamma^t}r_t]$. To encourage the adversary to systematically seek adverse outcomes, its modified value function for action selection is
\begin{equation}
\hat{Q}_{A}(s,a) = Q_A(s,a) + \lambda_A Var_{k}[Q^k_A(s,a)],
\label{eq:computeq2}
\end{equation}
where $\hat{Q}_A(s,a)$ is the modified Q function, $Q_A(s,a)$ is the original Q function, $Var_{k}[Q^k_A(s,a)]$  is the variance of the Q function across $k$ different models, and $\lambda_A$ is a  constant; the interaction between agents becomes a zero-sum game by setting $\lambda_A=\lambda_P$. The term $\lambda_A Var_{k}[Q^k_A(s,a)]$ is called the \emph{risk-seeking} term thereafter. The reward of $A$ is the negative of the environment reward $-r_t$, and its action space is the same as for the protagonist.
\begin{figure}[t]
\includegraphics[width=\linewidth]{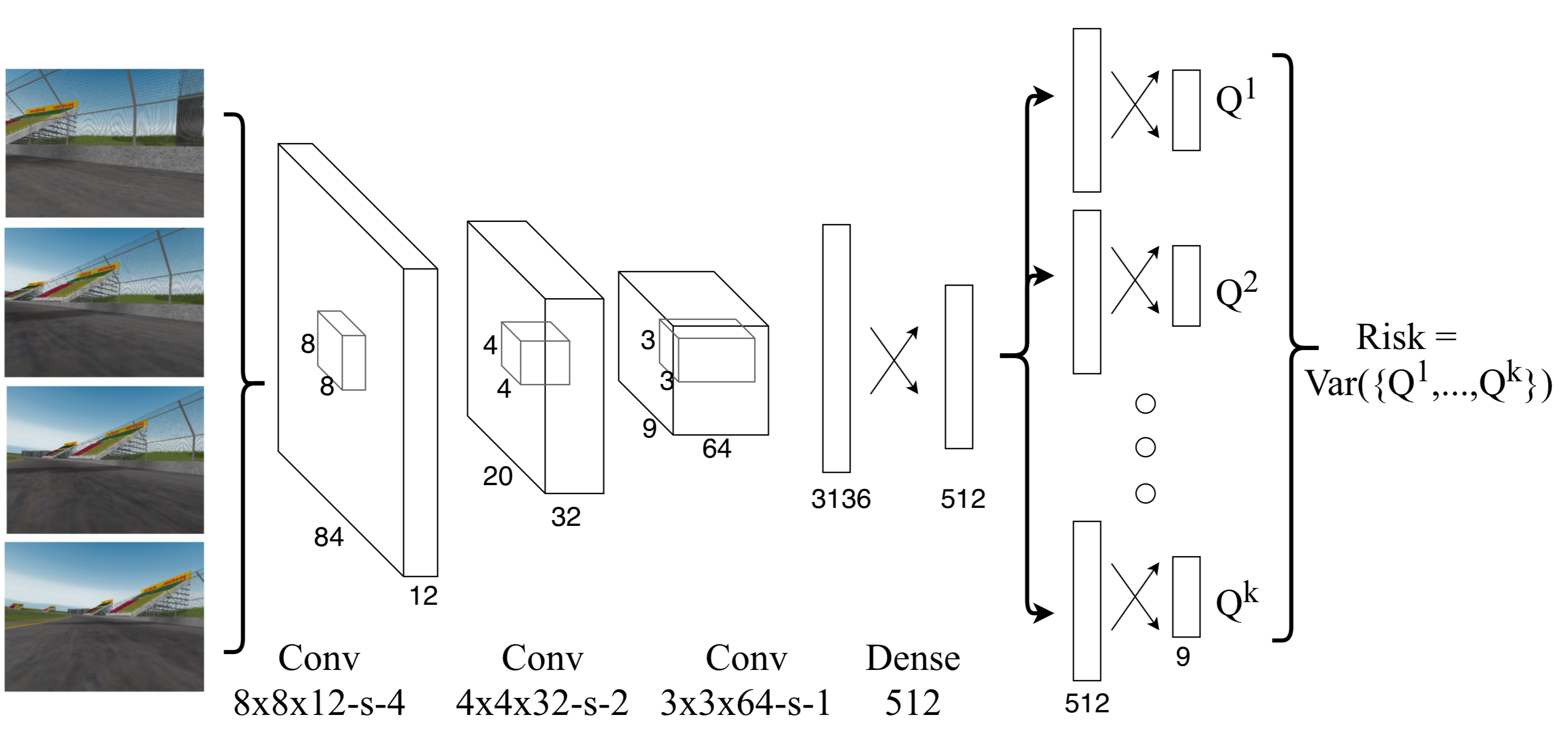}
\caption{
\footnotesize
Our neural network design. (Notation: ``s'' indicates stride for the convolutional weight kernel, and two crossing arrows indicate dense layers.) The input is a sequence of four stacked observations to form an $(84\times 84\times 12)$-dimensional input. It is passed through three convolutional layers to obtain a 3136-dimensional vector, which is then processed through a dense layer. (All activations are ReLus.) The resulting 512-dimensional vector is copied and passed to $k$ branches, which each process it through dense layers to obtain a state value vector $Q^i(s,\cdot)$. We apply the ensemble DQN framework for estimating the value function variance.
}
\label{fig:2}
\vspace*{-10pt}
\end{figure}

The necessity of having two agents working separately instead of jointly is to provide the adversary more power to create challenges for the protagonist. For example, in autonomous driving, a single risky action may not put the vehicle in a dangerous condition. In order to create a catastrophic event (e.g., a traffic accident) the adversary needs to be stronger. In our experiments (a vehicle controller with discrete control), the protagonist and adversary alternate full control of a vehicle, though our methods also apply to settings in~\cite{pinto2017robust}, where the action applied to the environment is a sum of contributions from the protagonist and the adversary.

% Daniel: I propose merging this with the next section.
%\subsection{Asymmetric Reward Design}
%To train risk averse agents, we propose an asymmetric reward function design such that good behavior receives small positive rewards and risky behavior receives \emph{very} negative rewards. We train multiple models to estimate the value of an action. The intuition is that having higher variance among the model estimates for action values is a good clue that an action might cause a catastrophe. See Section~\ref{ssec:sim_env} and Equation~\ref{eq:reward1} for our reward function.
%
% \begin{figure}[t]
% \centering
% \includegraphics[width=0.6\linewidth]{reward.png}
% \caption{Asymmetric Reward Design}
% \end{figure}

\subsection{Reward Design and Risk Modeling}

To train risk averse agents, we propose an asymmetric reward function design such that good behavior receives small positive rewards and risky behavior receives \emph{very} negative rewards. See Section~\ref{ssec:sim_env} and Equation~\ref{eq:reward1} for details.

% Daniel: this used to be in the old section, but we repeat it right here.
%We train multiple models to estimate the value of an action. The intuition is that having higher variance among the model estimates for action values is a good clue that an action might cause a catastrophe. 

The risk of an action can be modeled by estimating the variance of 
the value function across different models trained on different sets of data.
Inspired by~\cite{osband2016deep}, we estimate the variance of Q value functions
by training multiple Q value networks in parallel. 
Hereafter, we use $Q$ to denote
the entire Q value network, and use $Q^i$ to
denote the $i$-th head of the multi-heads
Q value network.\footnote{We use $Q$ and $Q^i$ to represent functions that could apply to either the protagonist or adversary. If it is necessary to distinguish among the two agents, we add the appropriate subscript of $P$ or $A$.} As shown in Figure~\ref{fig:2},
the network takes in input $s$, which consists of stacked frames of consecutive observations. It passes $s$ through three shared convolutional layers, followed by one (shared) dense layer. After this, the input is passed to $k$ different heads which perform one dense layer to obtain $k$ 
action-value outputs: $\{ Q^1(s, \cdotp), \ldots, Q^k(s, \cdotp) \}$. Defining 
the mean as $\tilde{Q}(s,a) = \frac{1}{k}\sum_{i=1}^kQ^i(s,a)$, the variance
of a single action $a$ is,
\begin{equation}
Var_k(Q(s,a)) = \frac{1}{k}\sum_{i=1}^k(Q^i(s,a)-\tilde{Q}(s,a))^2,
\label{eq:valuevar}
\end{equation}
where we use the $k$ subscripts to indicate variance over $k$ models, as in Equations~\ref{eq:computeq1} and~\ref{eq:computeq2}. The variance in Equation~\ref{eq:valuevar} measures risk, and our goal is for the protagonist and adversarial agents to select actions with low and high variance, respectively.

At training time, when we sample one action using the Q  values, we randomly choose one of $k$ heads from $Q^1$ to $Q^k$, and use this head throughout one episode to choose the action that will be applied by the agent. When updating Q functions, our algorithm (like DQN~\cite{mnih2015human}) samples a batch of data of size $B$ from the replay buffer $\{(s,a,s',r,done)_t\}_{t=1}^B$ which, for each data point, includes the state, action, next state, reward, and task completion signal. Then we sample a $k$-sized mask. Each mask value is sampled using a Poisson distribution (modeling a true Bootstrap sample with replacement) instead of the Bernoulli distribution in~\cite{osband2016deep} (sample without replacement).  
At test time, the mean value $\tilde{Q}(s,a)$ is used for selecting actions. 
\subsection{Risk Averse RARL}
% Daniel: this is not relevant to the discussion.
% We next consider the optimization problem given this unique two-player structure. As an example, we used Deep Q Learning (DQN) \cite{mnih2015human}
% as the basic reinforcement learning framework. In DQN, the agent keeps 
% a target value function $Q_{\theta^{*}}$ and an alternative value function $Q_{\theta}$, where $\theta^{*}$ and $\theta$ with parameters $\theta^*$ and $\theta$.\footnote{\bnote{Daniel: ``One batch of experience'' usually means a minibatch of (say) 32 examples, yet the notation here suggests just one ... why not do something like $\{(...)\}_{i=1}^B$? And this minibatch notation is different from the ``$r$'' and ``$done$'' that you have in the previous section.}} At every update step, the algorithm samples one batch of experience $(s,a,s',r(s,a))$ from the experience replay buffer, and perform gradient 
% updates using the standard Bellman update. 
% \begin{equation}
% \begin{split}
% \nabla_{\theta} = & \nabla_{\theta}\big[r(s,a)+\gamma\max_{a'}Q^{*}(s',a'|\theta^{*})
% -Q(s,a|\theta)\big]^2. 
% \end{split}
% \end{equation}
In our two-player framework, the agents take actions sequentially, not simultaneously: the protagonist takes $m$ steps, the adversary takes $n$ steps, and the cycle repeats. The 
experience of each agent is only visible to itself, which means each agent
changes the environment transition dynamics for another agent. The Q 
learning Bellman equation is modified to be compatible with this case.
Let the current and target value functions be $Q_P$ and $Q_P^*$ for the protagonist, and (respectively) $Q_A$ and $Q_A^*$ for the adversary. Given the
current state and action pair $(s_t,a_t)$, we denote actions executed by the protagonist as $a_{t}^P$ and actions taken by the adversary as $a_{t}^A$. The target value functions are $
Q_P(s_{t}^P, a_{t}^P) = \; r(s_{t}^P, a_{t}^P) + \sum_{i=1}^n\gamma^i r(s_{t+i}^A, a_{t+i}^A)
 + \gamma^{n+1}\max_{a}Q^*(s_{t+n+1}^P, a), $
and, similarly,$
Q_A(s_{t}^A, a_{t}^A) =  \; r(s_{t}^A, a_{t}^A) + \sum_{i=1}^m\gamma^i r(s_{t+i}^P, a_{t+i}^P) 
 + \gamma^{m+1}\max_{a}Q^*(s_{t+m+1}^A, a). $
%\begin{equation}
%\begin{split}
%Q_P(s_{t}^P, a_{t}^P) = & \; r(s_{t}^P, a_{t}^P) + \sum_{i=1}^n\gamma^i r(s_{t+i}^A, a_{t+i}^A)\\
%& + \gamma^{n+1}\max_{a}Q^*(s_{t+n+1}^P, a), \\
%Q_A(s_{t}^A, a_{t}^A) = & \; r(s_{t}^A, a_{t}^A) + \sum_{i=1}^m\gamma^i r(s_{t+i}^P, a_{t+i}^P) \\
% &+ \gamma^{m+1}\max_{a}Q^*(s_{t+m+1}^A, a). \\
%\end{split}
%\end{equation}
% the problem
% can be transformed into a single-agent problem and consider the other agent
% as part of the environment. The gradient updates would be the same for each
% agent. However, given the environmental reward $r$, the true reward for
% the protagonist and adversarial agent would be,
% \begin{equation}
% \begin{split}
% r_P & = r \\
% r_A & = -r \\
% \end{split}
% \end{equation}
% So that using the same framework the adversarial agent is effectively 
% minimizing the cumulative expected reward. 
To increase training stability for the protagonist, we designed a \emph{training schedule} $\Xi$ of the adversarial agent. For the first $\xi$ steps, only the protagonist agent takes actions. After that, for every $m$ steps taken by the protagonist, the adversary takes $n$ steps.  The reason for this training schedule design is that we observed
if the adversarial agent is added too early (e.g., right at the start), the protagonist is unable to attain any rewards. Thus, we let the protagonist undergo a sufficient amount of training steps to learn basic skills. The use of masks in updating Q value functions is similar to~\cite{osband2016deep}, where the mask
is a integer vector of size equal to
batch size times number of ensemble Q
networks, and is used to determine
which model is to be updated with the sample
batch.
Algorithm~\ref{alg1} describes our training algorithm. 

\begin{algorithm}[h]
 \KwResult{Protagonist Value Function $Q_{P}$; Adversarial Value Function $Q_{A}$.}
 \textbf{Input:} Training steps T; Environment $env$; Adversarial Action Schedule $\Xi$; Exploration rate $\epsilon$; Number of models $k$.\\
 \textbf{Initialize:} $Q_{P}^i$, $Q_{A}^i$ ($i=1,\cdots,k$);\ Replay Buffer $RB_{P}$, $RB_{A}$; Action choosing head $H_P$, $H_A\in[1,k]$; t = 0; Training frequency $f$; Poisson sample rate $q$\;
 \While{$t < T$}{
  Choose Agent $g$ from $\{A(\textit{Adversarial agent}), P(\textit{Protagonist agent})\}$ according to $\Xi$ \;
  Compute $\hat{Q}_g(s,a)$ according to \eqref{eq:computeq1} and \eqref{eq:computeq2} \;
  Select action according to $\hat{Q}_g(s,a)$ by applying $\epsilon$-greedy strategy \;
  Excute action and get $obs, reward, done$\;
  %obs, reward, done = $env.$step(action) \;
  $RB_g  = RB_g \cup \{(obs, reward,done)\}$\;
  %Generate mask from $Possion(\lambda)$\;
  \If{t \% f = 0}{
    Generate mask $M\in\mathbb{R}^k\sim Poisson(q)$\;
    Update $Q_{P}^i$ with $RB_P$ and $M_i$, $i=1,2,...,k$\;
    Update $Q_{A}^i$ with $RB_A$ and $M_i$, $i=1,2,...,k$\;
  }
  \If{done}{
  update $H_p$ and $H_a$ by randomly sampling integers from 1 to $k$ \;
  reset $env$;}
  t = t + 1;
 }
 \caption{Risk Averse RARL Training Algorithm}
 \label{alg1}
\end{algorithm}

\section{Experiments}\label{sec:experiments}

We evaluated models trained by RARARL on an autonomous driving environment, TORCS~\cite{wymann2000torcs}. Autonomous driving has been explored in recent contexts for policy learning and safety~\cite{multimodal_sensors_2017,gradient_free_2017,steering_bounds_2018} and is a good testbed for risk-averse reinforcement learning since it involves events (particularly crashes) that qualify as catastrophes.

% \begin{figure}[t]%
% \centering
% \includegraphics[width=0.31\linewidth]{frontview.png}
% \includegraphics[width=0.31\linewidth]{topview.png}
% \includegraphics[width=0.31\linewidth]{map.png}
% \caption{\footnotesize The TORCS environment. From left to right: front view, top view, map. The left view corresponds to an example $(84 \times 84 \times 3)$-shaped 3-channel RGB image that (when frame-stacked appropriately) is input to the policies for our agents.}
% \label{fig:simenv}
% \end{figure}

\iffalse
\begin{figure*}
\centering
\includegraphics[width=0.45\textwidth,height=0.4\textwidth]{test_with_adv5.png}
\includegraphics[width=0.45\textwidth,height=0.4\textwidth]{test_with_random_5.png}
\caption{Testing Under Adversarial Perturbations, with the protagonist taking $n_P$ step and adversarial agent taking
$n_A$ steps (left); Testing Under Random Perturbations, with the protagonist taking $n_P$ step and then randomly taking $n_A$ steps (right).For all models with the same
name label, we use the same model but tested them under different conditions. We
tested the each model for 10 episodes, and the curves' shade shows the standard
deviation across 10 episodes. }
\label{res3}
\end{figure*}
\fi

\subsection{Simulation Environment}\label{ssec:sim_env}
For experiments, we use the Michigan Speedway environment in TORCS~\cite{wymann2000torcs}, which is a round way racing track; see Figure~\ref{fig:adv_agent} for sample observations. The states are $(84\times 84 \times 3)$-dimensional RGB images. The vehicle can execute nine actions: 
(1) move left and accelerate, 
(2) move ahead and accelerate, 
(3) move right and accelerate,
(4) move left, 
(5) do nothing, 
(6) move right, 
(7) move right and decelerate,
(8) move ahead and decelerate, 
(9) move right and decelerate. 

We next define our asymmetric reward function. Let $v$ be the magnitude of the speed, $\alpha$ be the angle between the speed and road direction, $p$ be the distance of the vehicle to the center of the road, and $w$ be the road width. We additionally define two binary flags: $\mathbf{1}_{\rm st}$ and $\mathbf{1}_{\rm da}$, with $\mathbf{1}_{\rm st} = 1$ if the vehicle is stuck (and 0 otherwise) and $\mathbf{1}_{\rm da} = 1$ if the vehicle is damaged (and 0 otherwise). Letting $C = \left\lceil{\frac{\mathbf{1}_{\rm st} + \mathbf{1}_{\rm da}}{2}}\right\rceil$, the reward function is defined as:
{\small\begin{equation}\label{eq:reward1}
\begin{split}
r   = & \beta v \left(\cos(\alpha) - |\sin(\alpha)| - \frac{2p}{w}\right) (1-\mathbf{1}_{\rm st})(1-\mathbf{1}_{\rm da})  + r_{\rm cat} \cdot C %\left\lceil{\frac{\mathbf{1}_{\rm st} + \mathbf{1}_{\rm da}}{2}}\right\rceil
\end{split}
\end{equation}}
with the intuition being that $\cos(\alpha)$ encourages speed direction along the road direction, $|\sin(\alpha)|$
penalizes moving across the road, and $\frac{2p}{w}$ penalizes driving
on the side of the road. We set the \emph{catastrophe reward} as $r_{\rm cat} = -2.5$ and set $\beta = 0.025$ as a tunable constant which ensures that the magnitude of the non-catastrophe
reward is significantly less than that of the catastrophe reward. The catastrophe reward measures collisions, which are highly undesirable events to be avoided. We note that constants $\lambda_P=\lambda_A$ used to blend reward and variance terms in the risk-augmented Q-functions in Equations~\ref{eq:computeq1} and~\ref{eq:computeq2} were set to $0.1$.

We consider two additional reward functions to investigate in our experiments. The \emph{total progress reward} excludes the catastrophe reward:
\begin{equation}\label{eq:reward2}
r = \beta v \left(\cos(\alpha)-|\sin(\alpha)|-\frac{2p}{w}\right)
(1-\mathbf{1}_{\rm st})(1-\mathbf{1}_{\rm da}),
\end{equation}
and the \emph{pure progress reward} is defined as
\begin{equation}\label{eq:reward3}
    r=\beta v\Big(\cos(\alpha)-|\sin(\alpha)|\Big) (1-\mathbf{1}_{\rm st})(1-\mathbf{1}_{\rm da}).
\end{equation}
The total progress reward considers both moving along the road and across the road, and penalizes large distances to the center of the road, while the pure progress only measures the distance traveled by the vehicle, regardless of the vehicle's location. The latter can be a more realistic measure since vehicles do not always need to be at the center of the road.

\subsection{Baselines and Our Method}\label{ssec:experiments_legend}

All baselines are optimized using Adam~\cite{kingma2014adam} with learning rate 0.0001 and batch size 32. In all our ensemble DQN models, we trained with 10 heads since empirically that provided a reasonable balance between having enough models for variance estimation but not so much that training time would be overbearing. For each update, we sampled 5 models
using Poisson sampling with $q=0.03$ to generate the mask for updating Q value functions. We set the training frequency as 4, the 
target update frequency as 1000, and the replay buffer size as 100,000. For training DQN with an epsilon-greedy strategy, the $\epsilon$ decreased linearly
from 1 to 0.02 from step 10,000 to step 500,000. The time
point to add in perturbations is $\xi=550,000$ steps, and
for every $m=10$ steps taken by protagonist agent, the
random agent or adversary agent will take $n=1$ step.

\textbf{Vanilla DQN}. The purpose of comparing with vanilla DQN is to show that models trained in one environment may overfit to specific dynamics and fail to transfer to other environments, particularly those that involve random perturbations. We denote this as \texttt{\textbf{dqn}}. 

\textbf{Ensemble DQN}. Ensemble DQN tends to be more robust
than vanilla DQN. However,
without being trained on different dynamics, even Ensemble DQN may
not work well when there are adversarial attacks or simple random changes in the dynamics. We denote this as \texttt{\textbf{bsdqn}}.

\textbf{Ensemble DQN with Random Perturbations Without 
Risk Averse Term}. We train the protagonist and provide random perturbations according to the schedule $\Xi$. We do not include the variance guided exploration term here, so only the Q value function is used for choosing actions. The schedule $\Xi$ is the same as in our method. We denote this as \texttt{\textbf{bsdqnrand}}.

\textbf{Ensemble DQN with Random Perturbations With the Risk Averse Term}. We only train the 
protagonist agent and provide random perturbations according to the adversarial
training schedule $\Xi$. The protagonist selects action based on its 
Q value function and the risk averse term. We denote this as \texttt{\textbf{bsdqnrandriskaverse}}.

\textbf{Ensemble DQN with Adversarial Perturbation}. This is to compare our
model with \cite{pinto2017robust}. For a fair comparison, we also use
Ensemble DQN to train the policy while the variance term is not used as 
either risk-averse or risk-seeking term in either agents. We denote 
this as \texttt{\textbf{bsdqnadv}}.

\textbf{Our method}. In our method, we train both the protagonist and the adversary with Ensemble DQN. We include here the variance guided 
exploration term, so the Q function and its variance across 
different models will be used for action selection. The adversarial
perturbation is provided according to the adversarial training schedule $\Xi$.
We denote this as \texttt{\textbf{bsdqnadvriskaverse}}. 

% For \texttt{{Exp3, Exp4, Exp5}}, the adversarial perturbations or random perturbations
% schedule is shown in figure~\ref{value_diff}. The actions are taken as such, the protagonist
% will take $n_P=10$ steps, and then the adversarial agent takes $n_A$ steps. The ratio of 
% $n_A$ over $n_P$ is shown in the last figure in figure~\ref{value_diff}. This is referred as $\Xi$.  
\subsection{Evaluation}

\begin{figure*}[htb!]
\centering
\includegraphics[width=0.96\linewidth]{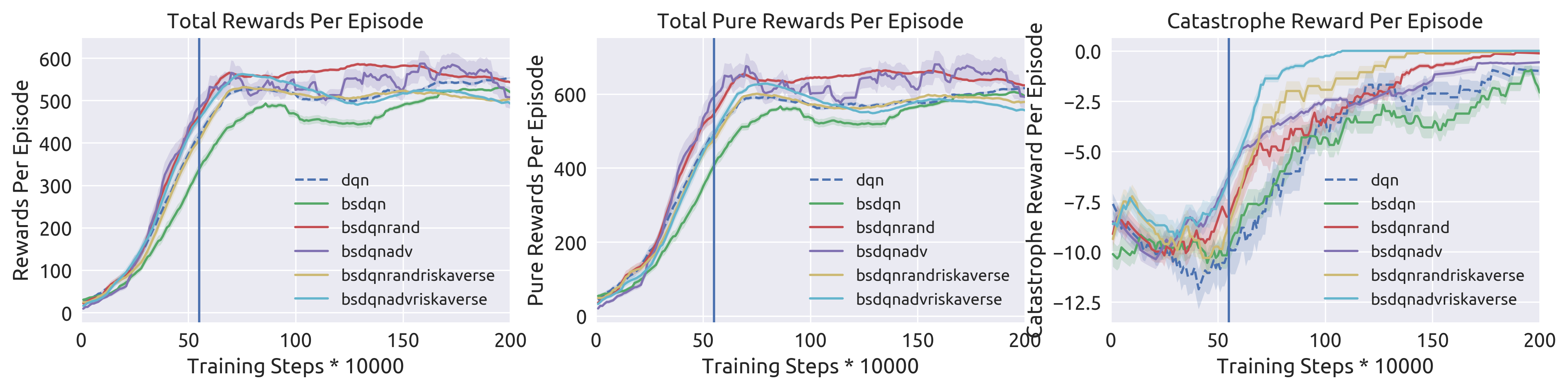}
\caption{
\footnotesize
Testing all models without attacks or perturbations. The reward is divided 
into distance related reward (left subplot), progress related reward (middle subplot). We also present results for catastrophe reward \emph{per episode} (right subplot). The blue vertical line
indicates the beginning of adding
perturbations during training. 
All legends follow the naming convention described in Section~\ref{ssec:experiments_legend}. %\bnote{Daniel: more details needed here, e.g., to address reviewer comments. Then in the next two figures you can refer to the caption here.} 
}
\label{res2}
\end{figure*}

\begin{figure*}[htb!]
\centering
\includegraphics[width=0.96\linewidth]{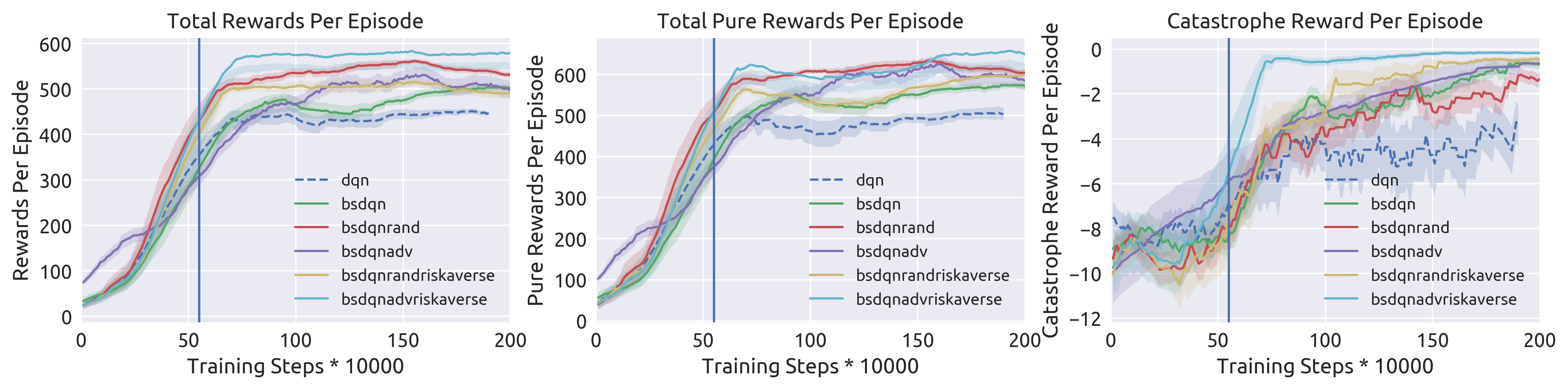}
\caption{
\footnotesize
Testing all models with random attacks. The three subplots follow the same convention as in Figure~\ref{res2}.}
\label{random}
\end{figure*}

To evaluate robustness of our trained models, we use the same trained models under different testing conditions, and evaluate using the previously-defined reward classes of total progress (Equation~\ref{eq:reward2}), pure progress (Equation~\ref{eq:reward3}), and additionally consider the reward of catastrophes.
We present three broad sets of results:
(1) \textbf{No perturbations}. (Figure~\ref{res2}) We tested all trained models from Section~\ref{ssec:experiments_legend} without perturbations.
(2)  \textbf{Random perturbations}. (Figure~\ref{random}) To evaluate the robustness of trained models in the presence of random environment perturbations, we benchmarked all trained models using random perturbations. For every 10 actions taken by the main agent, 1 was taken at random.
(3) \textbf{Adversarial Perturbations}. (Figure~\ref{res3}) To test the ability of our models to avoid catastrophes, which normally require deliberate, non-random perturbations, we test with a trained adversarial agent which took 1 action for every 10 taken by the protagonist.

% \textbf{Baseline4 DQN with Simple Adversarial}. We use the method in
% \cite{pinto2017robust} to introduce an adversarial agent to change limited
% numbers of parameters during training, and evaluate their performance with
% and without environment perturbations.

% We first present all results in testing all models. We then seek to answer
% the following questions. First, how does the strength of adversarial agent
% affect the robustness of trained models? Second, how does the risk seeking
% behavior and risk averse behavior of the adversarial agent and protagonist 
% agent affect the robustness of the trained models? 
% Fourth, how is adversarial perturbation different from random perturbation?

\begin{figure*}[htb!]
\centering
\includegraphics[width=0.96\linewidth]{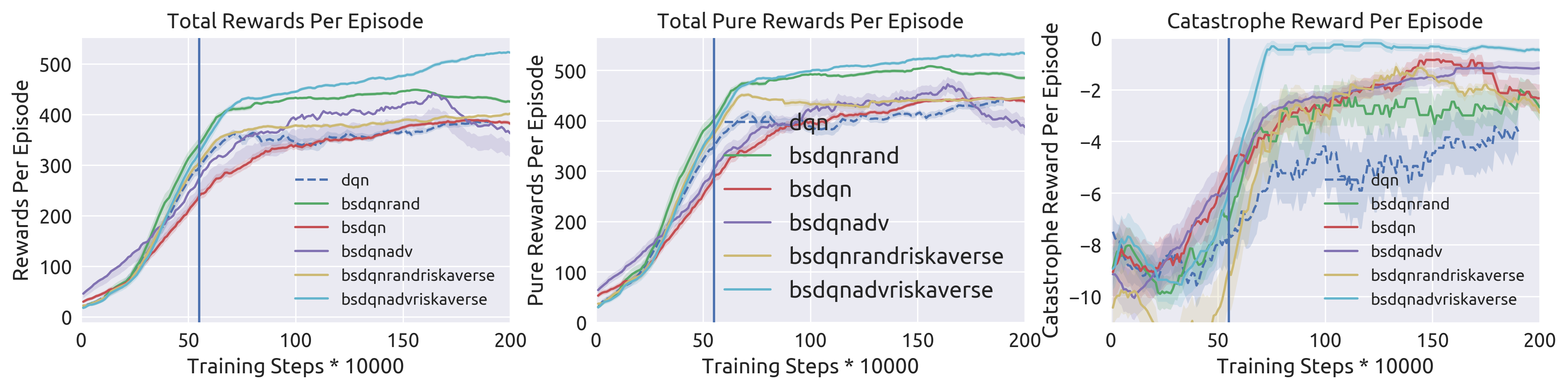}
\caption{
\footnotesize
Testing all models with adversarial attack. The three subplots follow the same convention as in Figure~\ref{res2}.}
\label{res3}
\end{figure*}

All subplots in Figures~\ref{res2},~\ref{random}, and~\ref{res3} include a vertical blue line at 0.55 million steps indicating when perturbations were first applied during training (if any). Before 0.55 million steps, we allow enough time for protagonist agents to be able to drive normally. We choose 0.55 million steps because the exploration rate decreases to 0.02 at 0.50 million steps, and we allow additional 50000 steps for learning to stabilize.

\begin{figure*}[htb!]
\centering
\includegraphics[width=\linewidth]{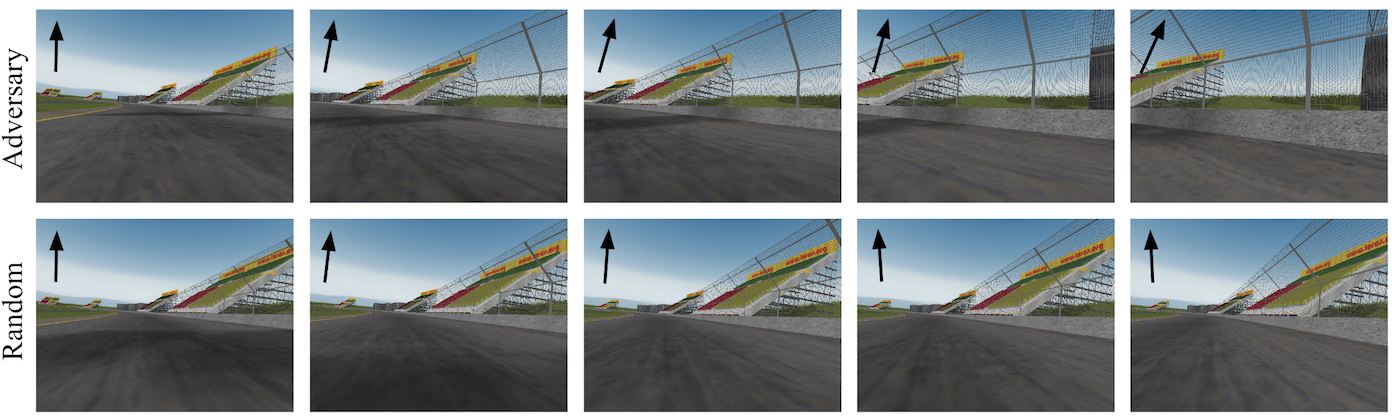}
\caption{
\footnotesize
Two representative (subsampled) sequences of states in TORCS for a trained protagonist, with either a trained adversary (top row) or random perturbations (bottom row) affecting the trajectory. The overlaid arrows in the upper left corners indicate the direction of the vehicle. The top row indicates that the trained adversary is able to force the protagonist to drive towards the right and into the wall (i.e., a catastrophe). Random perturbations cannot affect the protagonist's trajectory to the same extent because many steps of deliberate actions in one direction are needed to force a crash.}
\label{fig:adv_agent}
\end{figure*}

\textbf{Does adding adversarial agent's perturbation affect the robustness?} 
In Table~\ref{table1}, we compare the robustness of all models by their catastrophe rewards. The results indicate that adding perturbations improves a model's robustness, especially to adversarial attacks. 
DQN trained with random perturbations is not 
as robust as models trained with adversarial perturbations, since random perturbations are weaker than adversarial perturbations. 
\begin{table}[t]
\caption{Robustness of Models Measured by Average Best Catastrophe Reward Per Episode (Higher is better) }
\centering
\begin{tabular}{c|ccc}
\Xhline{2\arrayrulewidth}
Exp & Normal & Random Perturb & 
Adv. Perturb \\
\Xhline{2\arrayrulewidth}
dqn & -0.80 & -3.0 & -4.0 \\
bsdqn & -0.90 & -1.1 & -2.5 \\
bsdqnrand & -0.10 & -1.0 & -2.1  \\
bsdqnadv & -0.30 & -0.5 & -1.0 \\
bsdqnrandriskaverse & -0.09 & -0.4 & -2.0 \\
bsdqnadvriskaverse & \textbf{-0.08} & \textbf{-0.1} & \textbf{-0.1} \\
\Xhline{2\arrayrulewidth}
\end{tabular}
\label{table1}
\vspace*{-10pt}
\end{table}

\textbf{How does the risk term affect the robustness of the trained models? }
As shown in Figures~\ref{random} and~\ref{res3}, models trained with the risk term achieved better robustness
under both random and adversarial perturbations. We attribute this to the risk term
encouraging the adversary to aggressively explore regions with high risk while encouraging 
the opposite for the protagonist. 

\textbf{How do adversarial perturbations compare to random perturbations?}
A trained adversarial agent can enforce stronger perturbations than random perturbations. By comparing Figure~\ref{random} and Figure~\ref{res3}, we see that the adversarial perturbation provides stronger attacks, which causes the reward to be lower than with random  perturbations.

We also visualize an example of the differences between a trained adversary and random perturbations in Figure~\ref{fig:adv_agent}, which shows that a trained adversary can force the protagonist (a vanilla DQN model) to drive into a wall and crash.

%\textcolor{red}{Xinlei: I think
%maybe we remove figure 5 entirely and
%put these images in our supplementary
%videos since it really should be a video.}

\section{Conclusion}
We show that by introducing a notion of risk averse behavior, a protagonist agent trained with a learned adversary experiences substantially fewer catastrophic events during test-time rollouts as compared to agents trained without an adversary. Furthermore, a trained adversarial
agent is able to provide stronger perturbations than random perturbations and can provide a better training signal for the protagonist as compared to providing random perturbations. In future work, we will apply RARARL in other safety-critical domains, such as in surgical robotics.

{\small
\section*{Acknowledgments}
Xinlei Pan is supported by Berkeley Deep Drive. Daniel Seita is supported by a National Physical Science Consortium Fellowship.
}

\clearpage
\newpage
\bibliographystyle{IEEEtran}
\bibliography{example}  % .bib
\end{document}

\maketitle
\section{Visualizing Credits by Each Agent}
In order to make sure that the protagonist agent (denoted by $P$) and the 
adversarial agent (denoted by $A$) both learned to maximize their
respective reward, we can visualize the cumulative value difference in
each training episode. 
Since we have two players acting against each other, the advantage function
factorization is different here. First, define the discounted return 
$R_t = \sum_{t=\tau}^{\infty}\gamma^{t-\tau}r_t$, and recall the definition 
of $Q(s,a)$ function and the definition of value function $V(s)$
\begin{equation}
\begin{split}
Q(s,a)  = \mathbb{E}_{\pi}[R_t|s, a] \quad
V(s)  = \mathbb{E}_{a\sim\pi}[R_t|s]
\end{split} 
\end{equation}
The advantage function can be expressed as,
\begin{equation}
A(s, a) = Q(s, a) - V(s)
\end{equation}
The optimal Q function is defined as $Q^{*} = \max_{\pi}Q^{\pi}(s,a)$,
and it should satisfy the Bellman equation,
\begin{equation}
Q^{*}(s, a) = \mathbb{E}_{s'}[r+\gamma\max_{a'}Q^{*}(s',a')|s,a],
\end{equation}
and and the optimal value function $V^{*}(s)$ should be,
\begin{equation}
V^{*}(s) = \max_{a}Q^{*}(s,a).
\end{equation}
Therefore, we have,
\begin{equation}
Q^{*}(s,a) = \mathbb{E}_{s'}[r+\gamma V^{*}(s')|s,a].
\end{equation}
%Suppose the experience sequences are 
%$\{(s_1, a_{1A},r_1),
%(s_2, a_{2P}, r_2), (s_3, a_{3P}, r_3),(s_4, a_{4A}, r_4),\\
%(s_5, a_{5P}, r_5)\}$, where the subscript $A$ means that
%action is taken by the adversarial agent, and subscript $P$
%means that action is taken by the protagonist agent. 
We define the value function 
that depends on the step number $i$ as ,
\[  \tilde{V}^{*}(s_i) = \left\{
\begin{array}{ll}
      V^{*}(s_i) & \text{if action $i$ taken by protagonist agent} \\
     -V^{*}(s_i) & \text{if action $i$ taken by adversarial agent} \\
\end{array} 
\right. \]
We use an indicator function $I(i)$ to indicate which agent is actually taking that action, 
and if action at step $i$ is taken by protagonist agent, $I(i)=1$, else $I(i)=-1$. 
%Then we have the advantage of 
%taking action $a_i$ at state $s_i$ is,
%\begin{equation}
%\begin{split}
%A^{*}(s_1,a_1) & = Q^{*}_A(s_1,a_1) - V^{*}_A(s_1) \\
%& = \mathbb{E}_{s'}[r_1+\gamma V^{*}_P(s_2)|s,a]-V^{*}_A(s_1).
%A^{*}_i(s_i,a_i) & = I(i)Q^{*}_i(s_i,a_i) - I(i)V^{*}_i(s_i) \\
%& = \mathbb{E}_{s_{i+1}}[r_i+\gamma I(i+1) V^{*}_{i+1}(s_{i+1})|s_i,a_i]-I(i)V^{*}_i(s_i).
%\end{split}
%\end{equation}
We define the temporal value difference as,
\begin{equation}
TD(s_i,a_i) =\tilde{V}^{*}(s_{i+1}) - \tilde{V}^{*}(s_i)
\end{equation}
%For deterministic environment, the advantage is
%\begin{equation}
%A^{*}(s_i, a_i) = r_i+\gamma \tilde{V}^{*}(s_{i+1}) - \tilde{V}^{*}(s_{i}).
%\end{equation}
%If the consecutive two steps are taken by the same agent, 
%for example, step 2 and step 3, then we have,
%\begin{equation}
%A^{*}_{P}(s_2,a_2) = r_2+\gamma V^{*}_{P}(s_3)-V^{*}_P(s_2)
%\end{equation}
For an episode of length $T$, the 
cumulative temporal value difference got by protagonist agent is,
\begin{equation}
\begin{split}
TD_P =   \sum_{i=1}^{T-1} \frac{1+I(i)}{2} TD^{*}(s_i,a_i) 
=  \sum_{i=1}^{T-1} \frac{1+I(i)}{2}\big[\tilde{V}^{*}(s_{i+1})-\tilde{V}^{*}(s_i)\big],
\end{split}
\label{vdiff_pro}
\end{equation}
the cumulative temporal value difference got by the adversarial agent is,
\begin{equation}
\begin{split}
TD_A =   \sum_{i=1}^{T-1} \frac{1-I(i)}{2} TD^{*}(s_i,a_i)  
=   \sum_{i=1}^{T-1} \frac{1-I(i)}{2}\big[ \tilde{V}^{*}(s_{i+1})-\tilde{V}^{*}(s_i)\big],
\end{split}
\label{vdiff_adv}
\end{equation}
To compute the cumulative temporal value difference, we used the target
value function as $V^{*}(s_i)$.